\documentclass{article}

\PassOptionsToPackage{numbers, compress}{natbib}
\usepackage[final]{nips_2018}

\usepackage[utf8]{inputenc} % allow utf-8 input
\usepackage[T1]{fontenc}    % use 8-bit T1 fonts
\usepackage{hyperref}       % hyperlinks
\usepackage{url}            % simple URL typesetting
\usepackage{booktabs}       % professional-quality tables
\usepackage{amsfonts}       % blackboard math symbols
\usepackage{nicefrac}       % compact symbols for 1/2, etc.
\usepackage{microtype}      % microtypography
\usepackage{amssymb}
\usepackage{array}
\usepackage{graphicx}
\usepackage{subcaption}
\usepackage{booktabs}
\usepackage{url}
\usepackage{algorithm}
\usepackage{algpseudocode}
\usepackage{mathtools}
\usepackage{amsmath}
\usepackage{blindtext}

\title{Repetitive Motion Estimation Network: Recover cardiac and respiratory signal from thoracic imaging}

\author{
  Xiaoxiao Li \\
  Yale University\\
  \And
  Vivek Singh\\
  Siemens Medical Solutions USA Inc.\\
  \AND
  Yifan Wu\\
  Siemens Medical Solutions USA Inc.\\
  \And
  Klaus Kirchberg\\
  Siemens Medical Solutions USA Inc.\\
  \And
  James Duncan\\
  Yale University\\
  \And
  Ankur Kapoor\\
  Siemens Medical Solutions USA Inc.\\
}

\begin{document}

\maketitle

\begin{abstract}
Tracking organ motion is important in image-guided interventions, but motion annotations are not always easily available. Thus, we propose Repetitive Motion Estimation Network (RMEN) to recover cardiac and respiratory signals. It learns the spatio-temporal repetition patterns, embedding high dimensional motion manifolds to 1D vectors with partial motion phase boundary annotations. Compared with the best alternative models, our proposed RMEN significantly decreased the QRS peaks detection offsets by $59.3\%$. Results showed that RMEN could handle the irregular cardiac and respiratory motion cases. Repetitive motion patterns learned by RMEN were visualized and indicated in the feature maps. \footnote{Disclaimer: This feature is based on research, and is not commercially available. Due to regulatory reasons its future availability cannot be guaranteed}
\end{abstract}

\section{Introduction}
Cardiac interventions are often performed under the guidance of projective X-ray imaging on a beating heart. In such cardiac image-guided interventions, the organ motion can cause artifacts in the acquired images and lead to misalignment between the static guidance information that is overlaid on these images to guide the physician\cite{mcclelland2013respiratory}. Studies to recover the organ motion signal directly from imaging by estimating explicit motion models \cite{low2010application,shechter2006displacement} have shown that specialized motion monitors or implanted markers are not necessary for accurate motion recovery. Recently, deep learning has been well-developed for video pattern recognition \cite{karpathy2014large} and count the repetition times \cite{levy2015live} and shown better performance in representing motion patterns in natural videos. Different from natural videos, cardiac imaging does not always have obvious region of interests (ROIs). Thus the motion annotations are not easily available without extra monitors or markers. Given the phase boundaries of one repetitive motion, we proposed a deep convolutional network architecture. It automatically encodes the spatial features and learns temporal repetitive patterns, to embed high dimensional video data to lower dimensional 1D curve. The softened ECG signals were used as the pseudo targets to learn the repetitive motion manifolds embedding transformation function. Hereby, simple frequency filters enable to detect and separate cardiac and respiratory motion manifolds. In this study, we aim to recover repetitive cardiac and respiratory motion using deep learning model, without full annotations and hypothesis of the explicit motion model.

%*TODO: perhaps a bit more background to problem is needed. Note sure if NIPS community is knowlegeable about x-ray projections etc and complexity of problem * *more clinical driven  background?? -xiaoxiao*

\section{Methods}
\subsection{Input and Output Definition}
Mapping ECG to cardiac phase can be noisy. As a result, we only used phase boundaries (peaks) to train the network. We applied QRS wave detection algorithm \cite{ecgtool} to detect the peaks of the ECG signals for each cardiac video. The peak indices in X-ray fluoroscopic video $p_{X}$ were $p_{X} = \left\lfloor p_{E}\times f_{X}/{f_{E}}\right\rfloor$, given ECG sampling rate $f_{E}$ and X-ray fluoroscopic frame rate $f_{X}$ and the peak index in ECG signal $p_{E}$. We labeled the middle point of the two peaks was labeled as -1 and the peaks as 1. Assuming the underlying motion phase was smooth, we assigned the labels of the intermediate frames by linear interpolation. Then, we applied a sine transformation $\hat{x} = sin(x)$ to make smoother targets \cite{hinton2015distilling} as the outputs. Notably, the outputs signal were not periodic, which meant the intervals between two peaks varied. Inputs were cropped video sub-sequences of equal length.

\subsection{Repetitive Motion Estimation Network (RMEN) }
\label{netmethod}
\begin{figure}[htpb]
	\centering
    \includegraphics[width=12cm]{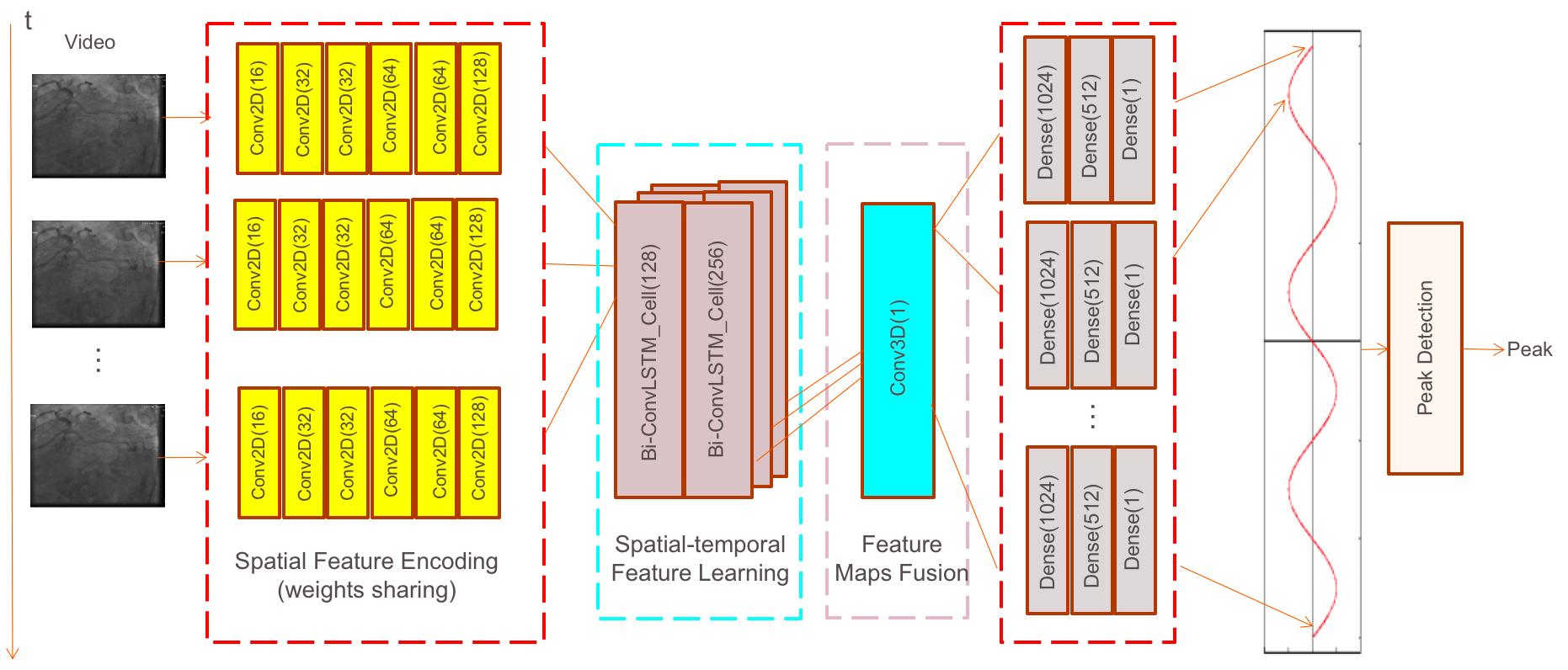}
    \caption{The architecture of Repetitive Repetitive Motion Estimation Network (RMEN) }
    \label{network}   
\end{figure}

The network architecture was shown in Fig \ref{network}. It has 5 parts (denoted in different colors).   The first weight sharing part was designed for encoding the spatial feature. Max pooling layers followed after the 1st, 3rd, and 5th Conv2D blocks. In the second part, the encoded frame features sequence then were passed to a 2-layer stacked ConvLSTM \cite{xingjian2015convolutional}. In the third part, a Conv3D layer convoluted across the channels for feature fusion. In the fourth part, the feature map encoded after Conv3D at each time point layer was flattened and passed to 3 fully connected layers, to regress the predicted phase. Mean squared error was used as the loss function. Dropout layers were added with ratio 0.5 after the first and the second dense layer to avoid overfitting. For each these frames, we had a prediction distribution $\{y^1,y^2,\cdots,y^Q\}$, where $y^q$ meant the prediction value when the frame is $q^{th}$ visited. We always kept the median values of the prediction distributions to generate the predicted phase curve. In the last part, we detected cardiac peaks from the filtered signals.

\subsection{Cardiac and Respiratory Signal Decomposition Algorithm}
Although, there is no ground truth label for respiratory signal learning, we believe RMEN can learn the repetitive motion manifolds embedding function. Given the prior knowledge - the cardiac rate is around $0.5 - 2$ $Hz$ and the respiration rate is around around $0.2 - 0.33$ Hz, we designed frequency filter to separate cardiac and respiratory signals from the predicted curve in \ref{netmethod}. Defining $H(\omega,\omega_0) = 0$ when $\omega < \omega_0$, else $H(\omega,\omega_0) = 1$, the zero phase shift band-pass filter : $G(\omega) =1-H(\omega,\omega_h) + H(\omega,\omega_l)$, was designed to filter out cardiac signal, where $\omega_l$ is low cutoff frequency ($\omega_l = 0.5$ Hz in our application), $\omega_h$ is high cutoff frequency ($\omega_h = 2$ Hz in our application). For filtering the breathing signal, a the zero phase shift low pass filter: $G(\omega) = 1 - H(\omega,\omega_b)$  was designed,where $\omega_b$ is respiration cutoff frequency ($\omega_b = 0.33$ Hz in our application). Naive peak detection algorithm \cite{findpeak} was applied on cardiac phase curve for peak detection.

\section{Experiments and Results}
\subsection{Cardiac/Respiratory Signal Recovery}
\begin{figure}[htpb]
	\centering
    \includegraphics[width=12cm]{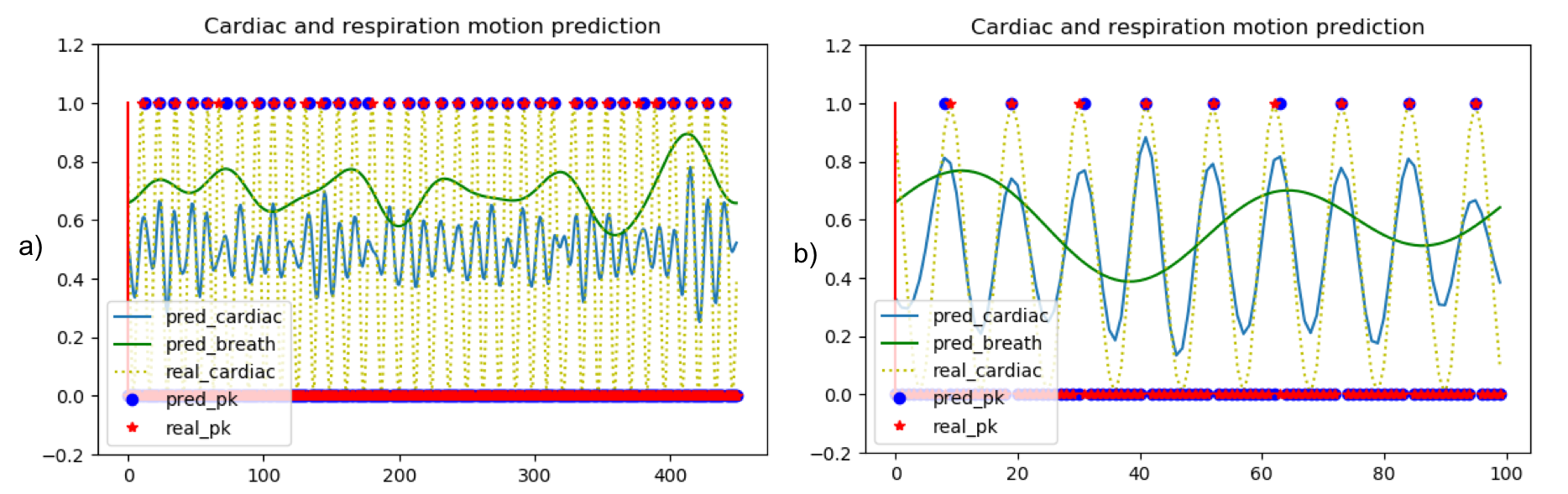}
    %\caption{Cardiac and respiratory signals decomposed from predictive curve: a) results from a long video with irregular cardiac and breathing motion; b) results from a short video with regular cardiac breathing motion. }
    \caption{Cardiac (blue line) and respiratory (green line) signals decomposed from predictive curve. }
    \label{signal}   
\end{figure}

We used two datasets, which covered different types of cardiac procedures to evaluate our model. The model was trained on the dataset \textit{A}, which contained 629 15fps fluoroscopic videos. We split 500 videos as the training set, 169 videos as the validation set for choosing the early stopping epoch. We tested the model on the dataset \textit{B}, which contained 329 15fps fluoroscopic videos.  Some example results are shown in \ref{signal}. RMEN also could handle the irregular cardiac and respiratory motion cases, such as video shifting, skipping cardiac cycle or breathing holding, etc.
\subsection{Cardiac Peak Detection Model Comparison}
We compared RMEN with naive LSTM and Support Vector Machine Regression (SVR), which took the 1D vector as input. The input to the naive LSTM models were flattened to a 1D vector.  We used principal component analysis (PCA) based dimensional reduction. The dimension-reduced vector accounted for more than 95\% of the variance of the original vector with 50 components. The best parameters for each alternative model was selected by validation. In addition, we compared with unsupervised learning methods, using the density change of the frame as signal. The model comparison results are shown in Table \ref{methodcompare}.

\begin{table}
  \caption{Peak Detection Results (window size = 20)}
  \centering
  \begin{tabular}{p{1cm}<{\centering} p{1cm}<{\centering} p{1.5cm}<{\centering} p{2cm}<{\centering} p{2cm}<{\centering}p{2cm}<{\centering}p{1cm}<{\centering}  } %{lcccccc}
    \toprule
    %\multicolumn{2}{c}{Part}                   \\
    %\cmidrule(r){1-2}
   Train  & Test  & Model & Offset/frame & True Negative & False Positive& Total   \\
    \midrule
 \textit{A} & \textit{B}&  RMEN & \textbf{0.88} & \textbf{3}& \textbf{27} & 1703 \\
   \textit{A}  & \textit{B} & PCA+LSTM & 2.16& 9& 72 & 1703\\
   \textit{A}  & \textit{B} & PCA+SVR& 2.67 &5& 80 & 1703\\
   N/A & \textit{B}& DensityFlow& 3.7& 2& 96 & 1703 \\
    \bottomrule
  \end{tabular}
  \label{methodcompare}
\end{table}
\vspace{-4mm}
\subsection{Visualizing Repetitive Patterns}
\begin{figure}[htpb]
	\centering
    \includegraphics[width=14cm]{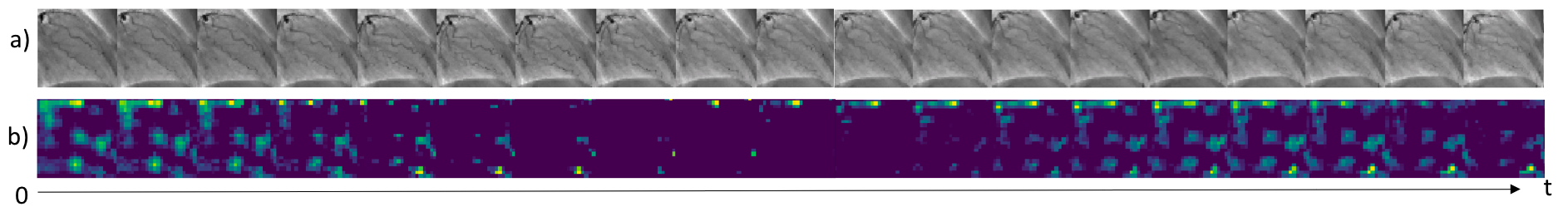}
    \caption{a): video sequence; and b)corresponding repetitive feature map of Conv3D layer }
    \label{pattern}   
\end{figure}

In order to understand what RMEN learned, we examined the feature maps of the Conv3D layers (Figure \ref{pattern}). The results indicated repetitive patterns over time, which highlighted the change of coronaries (related to cardiac motion) and diaphragm (related to respiratory motion).
\vspace{-4mm}
\section{Conclusion}
Our proposed deep learning pipeline significantly outperforms the alternative models in motion phase boundary detection. We found the repetitive patterns of cardiac and respiratory motion from the feature maps. The proposed RMEN embedded the repetitive motion manifolds to 1D signal, hence even the motions without annotation could be simply separated by frequency filtering. The proposed methods can be applied to a variety of intervention applications, such as dynamic coronary roadmapping and cardiac pacemaker tracking. It offers a solution to uncover repetitive motions, as long as the motion is smooth and partial annotations of motion boundaries are available. 

\bibliographystyle{ieeetr}
\bibliography{nipsphase}

\begin{thebibliography}{1}

\bibitem{mcclelland2013respiratory}
J.~R. McClelland, D.~J. Hawkes, T.~Schaeffter, and A.~P. King, ``Respiratory
  motion models: A review,'' {\em Medical image analysis}, vol.~17, no.~1,
  pp.~19--42, 2013.

\bibitem{low2010application}
D.~A. Low, T.~Zhao, B.~White, D.~Yang, S.~Mutic, C.~E. Noel, J.~D. Bradley,
  P.~J. Parikh, and W.~Lu, ``Application of the continuity equation to a
  breathing motion model,'' {\em Medical physics}, vol.~37, no.~3,
  pp.~1360--1364, 2010.

\bibitem{shechter2006displacement}
G.~Shechter, J.~R. Resar, and E.~R. McVeigh, ``Displacement and velocity of the
  coronary arteries: cardiac and respiratory motion,'' {\em IEEE Transactions
  on Medical Imaging}, vol.~25, no.~3, p.~369, 2006.

\bibitem{karpathy2014large}
A.~Karpathy, G.~Toderici, S.~Shetty, T.~Leung, R.~Sukthankar, and L.~Fei-Fei,
  ``Large-scale video classification with convolutional neural networks,'' in
  {\em Proceedings of the IEEE conference on Computer Vision and Pattern
  Recognition}, pp.~1725--1732, 2014.

\bibitem{levy2015live}
O.~Levy and L.~Wolf, ``Live repetition counting,'' in {\em Proceedings of the
  IEEE International Conference on Computer Vision}, pp.~3020--3028, 2015.

\bibitem{ecgtool}
C.~Carreiras, A.~P. Alves, A.~Louren\c{c}o, F.~Canento, H.~Silva, A.~Fred, {\em
  et~al.}, ``{BioSPPy}: Biosignal processing in {Python},'' 2015--.
\newblock [Online; accessed <today>].

\bibitem{hinton2015distilling}
G.~Hinton, O.~Vinyals, and J.~Dean, ``Distilling the knowledge in a neural
  network,'' {\em arXiv preprint arXiv:1503.02531}, 2015.

\bibitem{xingjian2015convolutional}
S.~Xingjian, Z.~Chen, H.~Wang, D.-Y. Yeung, W.-K. Wong, and W.-c. Woo,
  ``Convolutional lstm network: A machine learning approach for precipitation
  nowcasting,'' in {\em Advances in neural information processing systems},
  pp.~802--810, 2015.

\bibitem{findpeak}
E.~Jones, T.~Oliphant, P.~Peterson, {\em et~al.}, ``{SciPy}: Open source
  scientific tools for {Python},'' 2001--.

\end{thebibliography}
\end{document}